\pdfoutput=1

\documentclass[11pt]{article}

\usepackage[]{ACL2023}

\usepackage{times}
\usepackage{latexsym}

\usepackage[T1]{fontenc}

\usepackage[utf8]{inputenc}

\usepackage{microtype}

\usepackage{inconsolata}

\usepackage{amsmath, mathtools}
\makeatletter
\def\maketag@@@#1{\hbox{\m@th\normalfont\normalsize#1}}
\makeatother

\usepackage{amssymb}
\usepackage{tikz}
\usepackage{cancel}
\newcommand{\mathboldface}[1]{\boldsymbol{#1}}
\newcommand{\bm}[1]{\mathboldface{#1}}
\usepackage[most]{tcolorbox}
\tcbset{on line, 
        boxsep=4pt, left=0pt,right=0pt,top=0pt,bottom=0pt,
        colframe=white,colback=gray!15,  
        highlight math style={enhanced}
        }

\usepackage{subfig}
\usepackage{pgfplots}
\usepgfplotslibrary{fillbetween}
\usetikzlibrary{patterns}
\usetikzlibrary{pgfplots.groupplots}
\pgfplotsset{compat=1.17}
\usepackage{lscape}
\usepackage{multirow}
\usepackage{dblfloatfix}
\usepackage{footnote}
\usepackage{graphicx}
\usepackage{tikz}
\usepackage{booktabs, tabularx}
\usepackage{amssymb}
\usepackage{xcolor}

\newcommand\blfootnote[1]{%
  \begingroup
  \renewcommand\thefootnote{}\footnote{#1}%
  \addtocounter{footnote}{-1}%
  \endgroup
}

\makeatletter
\newcommand*{\encircled}[1]{\relax\ifmmode\mathpalette\@encircled@math{#1}\else\@encircled{#1}\fi}
\newcommand*{\@encircled@math}[2]{\@encircled{$\m@th#1#2$}}
\newcommand*{\@encircled}[1]{%
  \tikz[baseline,anchor=base]{\node[draw,circle,outer sep=0pt,inner sep=.2ex] {#1};}}
\makeatother

\title{DecompX: Explaining Transformers Decisions \\by Propagating Token Decomposition}

\author{
    Ali Modarressi$^{1,2\star}$ ~ Mohsen Fayyaz$^{3\star}$ ~ Ehsan Aghazadeh$^{3}$ \\
    \textbf{Yadollah Yaghoobzadeh$^{3}$}  ~ \textbf{Mohammad Taher Pilehvar$^{4}$} \\
    $^1$ Center for Information and Language Processing, LMU Munich, Germany \\
    $^2$ Munich Center for Machine Learning (MCML), Germany ~
    $^3$ University of Tehran, Iran \\
    $^4$ Tehran Institute for Advanced Studies, Khatam University, Iran \\
    \texttt{amodaresi@cis.lmu.de} ~
    \texttt{mohsen.fayyaz77@ut.ac.ir} ~
    \texttt{eaghazade1998@ut.ac.ir} \\
    \texttt{y.yaghoobzadeh@ut.ac.ir} ~
    \texttt{mp792@cam.ac.uk}
}

\begin{document}
\maketitle
\begin{abstract}
An emerging solution for explaining Transformer-based models is to use vector-based analysis on how the representations are formed. However, providing a faithful vector-based explanation for a multi-layer model could be challenging in three aspects: (1) Incorporating all components into the analysis, (2) Aggregating the layer dynamics to determine the information flow and mixture throughout the entire model, and (3) Identifying the connection between the vector-based analysis and the model's predictions. 
In this paper, we present \emph{DecompX} to tackle these challenges. 
DecompX is based on the construction of decomposed token representations and their successive propagation throughout the model without mixing them in between layers.
Additionally, our proposal provides multiple advantages over existing solutions for its inclusion of all encoder components (especially nonlinear feed-forward networks) and the classification head. The former allows acquiring precise vectors while the latter transforms the decomposition into meaningful prediction-based values, eliminating the need for norm- or summation-based vector aggregation.
According to the standard faithfulness evaluations, DecompX consistently outperforms existing gradient-based and vector-based approaches on various datasets.
Our code is available at \href{https://github.com/mohsenfayyaz/DecompX}{github.com/mohsenfayyaz/DecompX}.

\blfootnote{$^\star$ Equal contribution.}
\end{abstract}

\section{Introduction}
 \newcommand\qualclip{40}
\begin{figure}[t]
\centering
    \subfloat{
        \includegraphics[height=0.16\textheight, trim=0 0 0 0, clip] {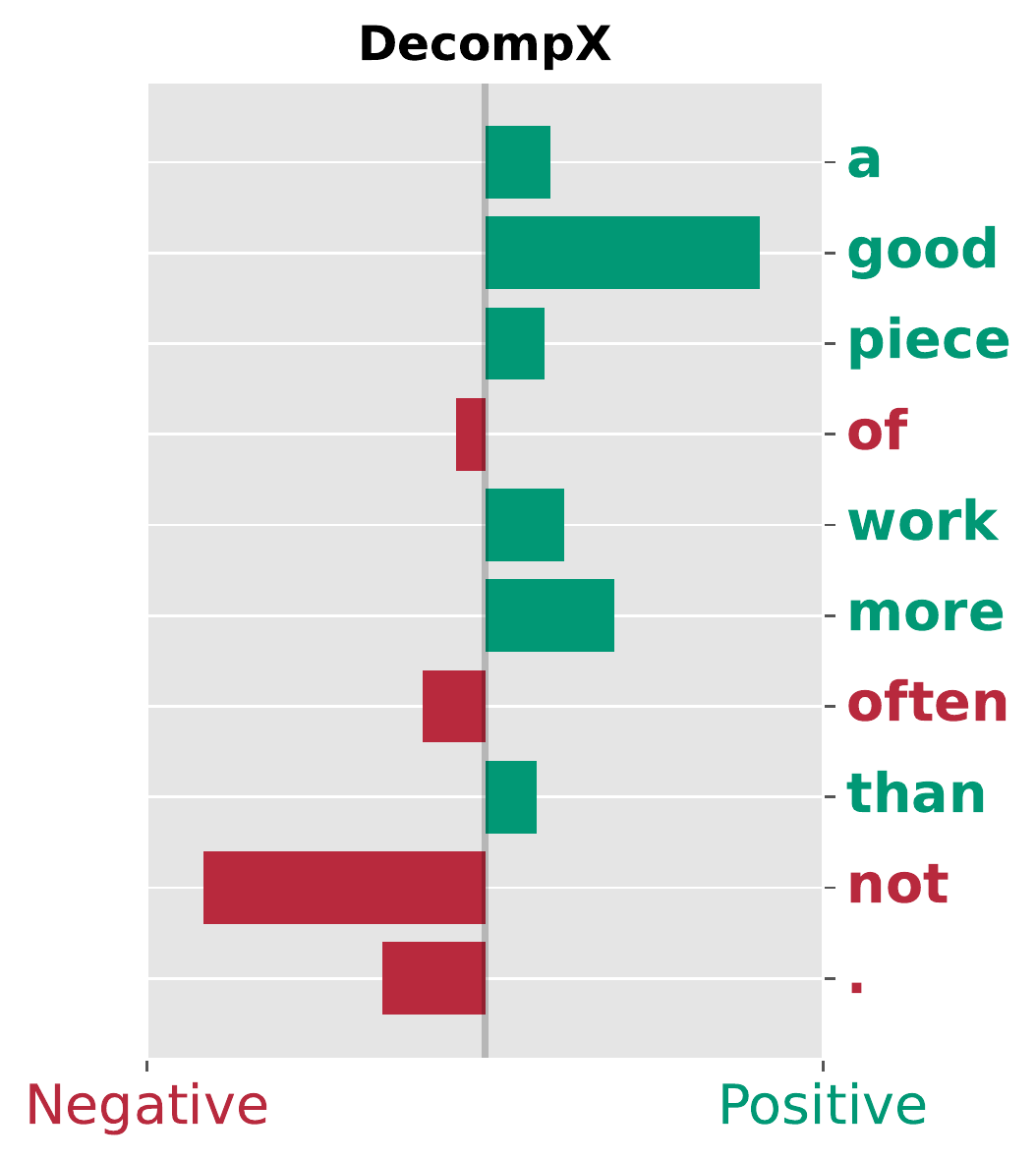}
    }
    \subfloat{
        \includegraphics[height=0.16\textheight, trim=5 0 70 0, clip] {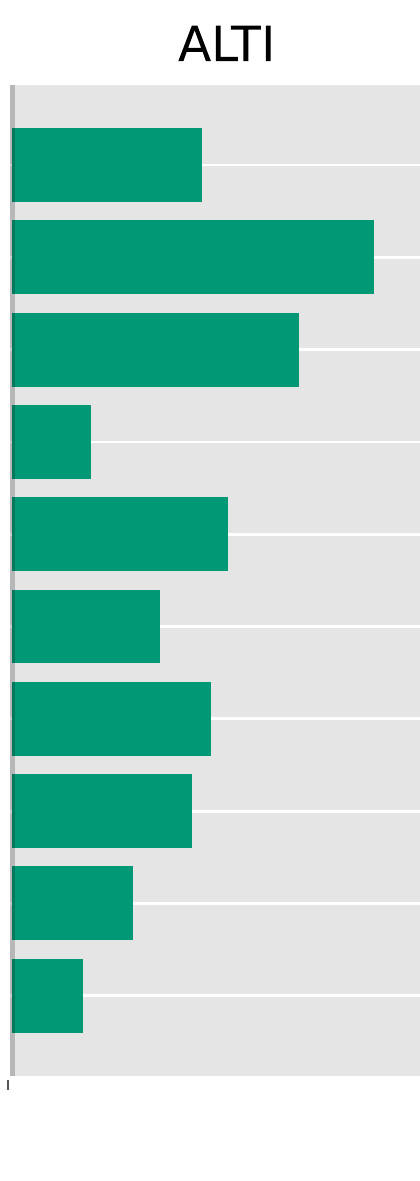}
    }
    \subfloat{
        \includegraphics[height=0.16\textheight, trim=5 0 0 0, clip] {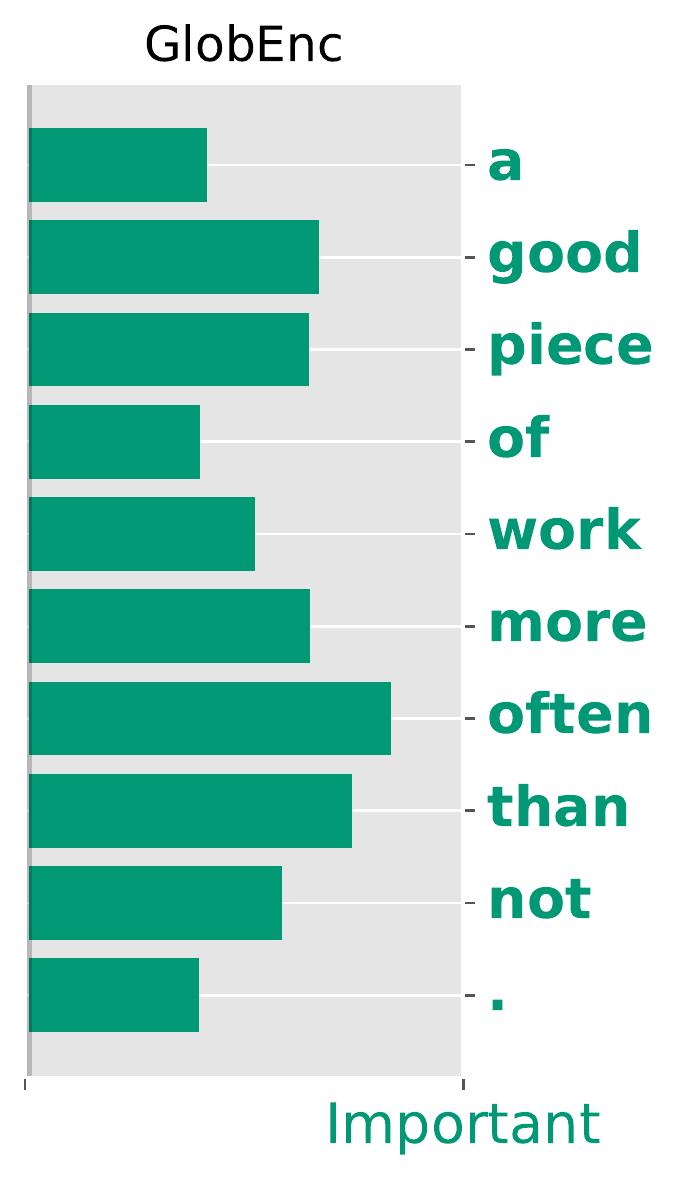}
    }
    \caption{
    The explanation of our method (DecompX) compared with GlobEnc and ALTI for fine-tuned BERT on SST2 dataset (sentiment analysis). Our method is able to quantify positive or negative attribution of each token as well as being more accurate.
    }
    \label{fig:examples}
\end{figure}

 While Transformer-based models have demonstrated significant performance, their black-box nature necessitates the development of explanation methods for understanding these models' decisions \cite{serrano-smith-2019-attention, bastings-filippova-2020-elephant, towards-faithful-survey-2022}.
On the one hand, researchers have adapted \emph{gradient-based} methods from computer vision to NLP \cite{li-etal-2016-visualizing, explain-explanations-bert-2021}. On the other hand, many have attempted to explain the decisions based on the components inside the Transformers architecture (\emph{vector-based} methods).
Recently, the latter has shown to be more promising than the former in terms of faithfulness \cite{ferrando-etal-2022-measuring}. 

Therefore, we focus on the vector-based methods which require an accurate estimation of (i) the mixture of tokens in each layer (\emph{local-level} analysis), and (ii) the flow of attention throughout multiple layers (\emph{global-level} analysis) \cite{pascual-etal-2021-telling-full-story}.
Some of the existing local analysis methods include raw attention weights \cite{clark-etal-2019-bert-look}, effective attentions \cite{Brunner2020On}, and vector norms \cite{kobayashi-etal-2020-attention-norm, kobayashi-etal-2021-incorporating-residual}, which all attempt to explain how a single layer combines its input representations.
Besides, to compute the global impact of the inputs on the outputs, the local behavior of all layers must be aggregated. \emph{Attention rollout} and \emph{attention flow} were the initial approaches for recursively aggregating the raw attention maps in each layer \cite{abnar-zuidema-2020-quantifying}. 
By employing rollout, GlobEnc \cite{modarressi-etal-2022-globenc} and ALTI \cite{ferrando-etal-2022-measuring} significantly improved on previous work by substituting norm-based local methods \cite{kobayashi-etal-2021-incorporating-residual} for raw attentions.
Despite their advancements, these vector-based methods still have three major limitations: (1) they ignore the encoder layer's Feed-Forward Network (FFN) because of its non-linearities, (2) they use rollout, which produces inaccurate results because it requires scalar local attributions rather than decomposed vectors which causes information loss, and (3) they do not take the classification head into account.

In an attempt to address all three limitations, in this paper, we introduce \emph{DecompX}. Instead of employing rollout to aggregate local attributions, DecompX propagates the locally decomposed vectors throughout the layers to build a global decomposition. Since decomposition vectors propagate along the same path as the original representations, they accurately represent the inner workings of the entire model. Furthermore, we incorporate the FFNs into the analysis by proposing a solution for the non-linearities. The FFN workaround, as well as the decomposition, enable us to also propagate through the classification head, yielding per predicted label explanations. 
Unlike existing techniques that provide absolute importance, this per-label explanation indicates the extent to which each individual token has contributed towards or against a specific label prediction (Figure~\ref{fig:examples}).

We conduct a comprehensive faithfulness evaluation over various datasets and models, that verifies how the novel aspects of our methodology contribute to more accurate explanations. Ultimately, our results demonstrate that DecompX consistently outperforms existing well-known gradient- and vector-based methods by a significant margin.

\begin{figure*}[t!]
\centering
    \subfloat{
        \includegraphics[width=0.95\textwidth, trim=0 120 0 0, clip] {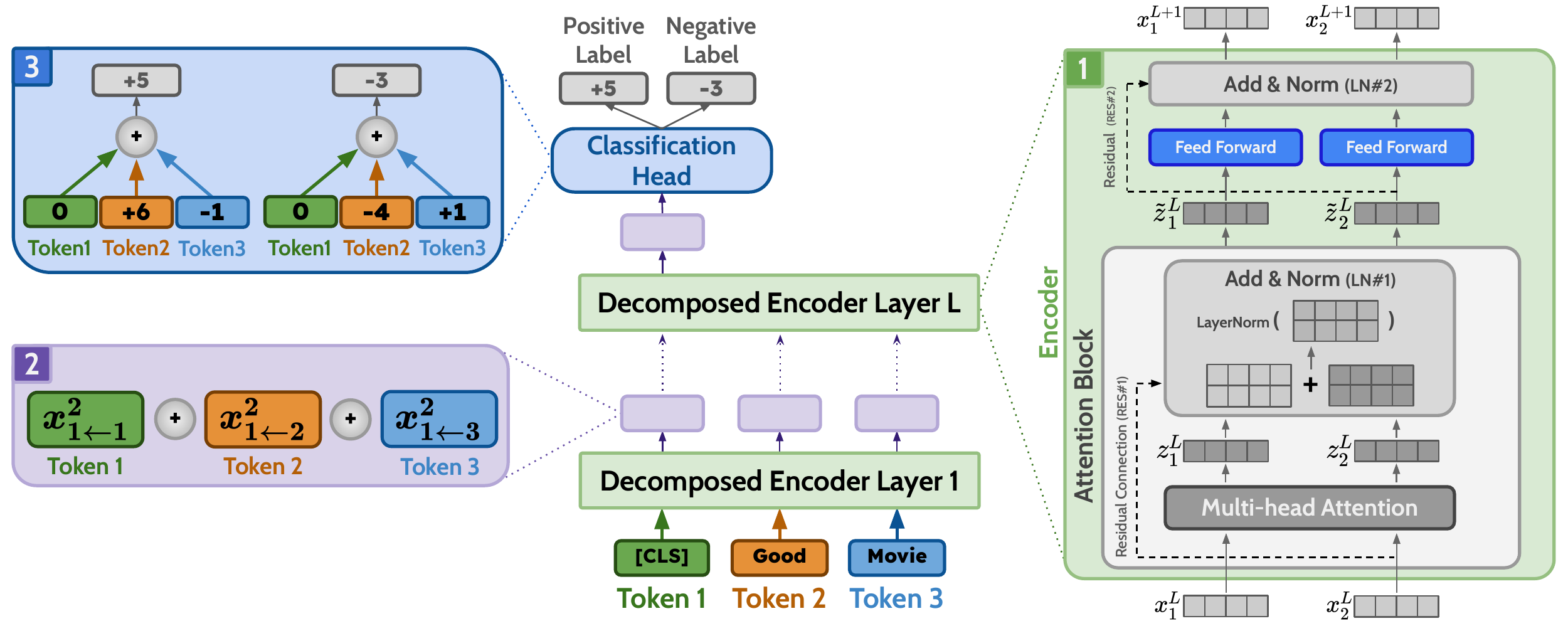}
    }
    \caption{
    The overall workflow of DecompX. The contributions include: \textcolor[HTML]{6AA84F}{\textbf{(1)}} incorporating all components in the encoder layer, especially the non-linear feed-forward networks; \textcolor[HTML]{7A64B2}{\textbf{(2)}} propagating the decomposed token representations through layers which prevents them from being mixed; and \textcolor[HTML]{3C78D8}{\textbf{(3)}} passing the decomposed vectors through the classification head, acquiring the exact positive/negative effect of each input token on individual output classes. 
    }
    \label{fig:decompx_diagram}
\end{figure*}

\section{Related Work}
Vector-based analysis has been sparked by the motivation that attention weights alone are insufficient and misleading to explain the model's decisions \cite{serrano-smith-2019-attention, jain-wallace-2019-attention}.
One limitation was that it neglects the self-attention value vectors multiplied by the attention weights. \citet{kobayashi-etal-2020-attention-norm} addressed it by using the norm of the weighted value vectors as a measure of inter-token attribution.
Their work could be regarded as one of the first attempts at Transformer decomposition. 
They expanded their analysis from the self-attention layer to the entire attention block and found that residual connections are crucial to the information flow in the encoder layer \cite{kobayashi-etal-2021-incorporating-residual}.

However, to be able to explain the multilayer dynamics, one needs to aggregate the local analysis into global by considering the attribution mixture across layers.
\citet{abnar-zuidema-2020-quantifying} introduce the attention rollout and flow methods, which aggregate multilayer attention weights to create an overall attribution map.
Nevertheless, the method did not result in accurate maps as it was based on an aggregation of attention weights only.
\emph{GlobEnc} \cite{modarressi-etal-2022-globenc} and \emph{ALTI} \cite{ferrando-etal-2022-measuring} improved this by incorporating decomposition at the local level and then aggregating the resulting vectors-norms with rollout to build global level explanations.
At the local level, GlobEnc extended \citet{kobayashi-etal-2021-incorporating-residual} by incorporating the second Residual connection and LayerNormalization layer after the attention block.
GlobEnc utilizes the L2-norm of the decomposed vectors as an attribution measure; however, \citet{ferrando-etal-2022-measuring} demonstrate that the reduced anisotropy of the local decomposition makes L2-norms an unreliable metric. 
Accordingly, they develop a scoring metric based on the L1-distances between the decomposed vectors and the output of the attention block. The final outcome after applying rollout, referred to as ALTI, showed improvements in both the attention-based and norm-based scores. 

Despite continuous improvement, all these methods suffer from three main shortcomings.
They all omitted the classification head, which plays a significant role in the output of the model. In addition, they only evaluate linear components for their decomposition, despite the fact that the FFN plays a significant role in the operation of the model \cite{geva-etal-2021-transformer, geva-etal-2022-transformer}. Nonetheless, the most important weakness in their analysis is the use of rollout for multi-layer aggregation.

Rollout assumes that the only required information for computing the global flow is a set of scalar cross-token attributions. 
Nevertheless, this simplifying assumption ignores that each decomposed vector represents the multi-dimensional impact of its inputs. Therefore, losing information is inevitable when reducing these complex vectors into one cross-token weight.
On the contrary, by keeping and propagating the decomposed vectors in DecompX, any transformation applied to the representations can be traced back to the input tokens without information loss.

\paragraph{Gradient-based methods.} One might consider gradient-based explanation methods as a workaround to the three issues stated above. Methods such as vanilla gradients \cite{Simonyan2014DeepIC}, GradientXInput \cite{kindermans2016investigating}, and Integrated gradients \cite{pmlr-v70-sundararajan17a} all rely on the gradients of the prediction score of the model w.r.t. the input embeddings. 
To convert the gradient vectors into scalar per-token importance, various reduction methods such as L1-norm \cite{li-etal-2016-visualizing}, L2-norm \cite{poerner-etal-2018-evaluating}, and mean \cite{atanasova-etal-2020-diagnostic, pezeshkpour-etal-2022-combining} have been employed. Nonetheless, \citet{bastings-etal-2022-will} evaluations showed that none of them is consistently better than the other.
Furthermore, adversarial analysis and sanity checks both have raised doubts about gradient-based methods' trustworthiness \cite{wang-etal-2020-gradient, adebayo2018sanity, Kindermans2019}.

\paragraph{Perturbation-based methods.} 
Another set of interpretability methods, broadly classified as perturbation-based methods, encompasses widely recognized approaches such as LIME \cite{ribeiromodel} and SHAP \cite{shapley1953value}. 
However, these were excluded from our choice of comparison techniques, primarily due to their documented inefficiencies and reliability issues as highlighted by \citet{atanasova-etal-2020-diagnostic}. 
We follow recent work \cite{ferrando-etal-2022-measuring,mohebbi-etal-2023-quantifying} and mainly compare against gradient-based methods which have consistently proven to be more faithful than perturbation-based methods. 

\citet{mohebbi-etal-2023-quantifying} recently presented a method called \emph{Value zeroing} to measure the extent of context mixing in encoder layers. Their approach involves setting the value representation of each token to zero in each layer and then calculating attribution scores by comparing the cosine distances with the original representations. 
Although they focused on local-level faithfulness, their global experiment has clear drawbacks due to its reliance on rollout aggregation and naive evaluation metric (cf. \ref{sec:roberta-results}).

\section{Methodology}

Based on the vector-based approaches of \citet{kobayashi-etal-2021-incorporating-residual} and \citet{modarressi-etal-2022-globenc}, we propose \emph{decomposing} token representations into their constituent vectors. 
Consider decomposing the $i^{th}$ token representation in layer $\ell \in \{0,1,2,...,L,L+1\}$\footnote{$\ell=0$ is the input embedding layer and $\ell=L+1$ is the classification head over the last encoder layer.}, i.e., $\bm{x}^\ell_i \in \{\bm{x}^\ell_1, \bm{x}^\ell_2, ..., \bm{x}^\ell_N\}$, into elemental vectors attributable to each of the $N$ input tokens:
\vspace{-1.0ex}
\begin{equation}
    \bm{x}^\ell_i=\sum_{k=1}^N\bm{x}^{\ell}_{i\Leftarrow k}
    \label{eq:decomp}
\vspace{-1.0ex}
\end{equation}
According to this decomposition, we can compute the norm of the attribution vector of the $k^{\text{th}}$ input ($\bm{x}^{\ell}_{i\Leftarrow k}$) to quantify its total attribution to $\bm{x}^\ell_i$. 
The main challenge of this decomposition, however, is how we could obtain the attribution vectors in accordance with the internal dynamics of the model. 

As shown in Figure \ref{fig:decompx_diagram}, in the first encoder layer, the first set of decomposed attribution vectors can be computed as $\bm{x}^2_{i\Leftarrow k}$.\footnote{As $\bm{x}$ denotes the inputs, the output decomposition of the first layer is the input of the second layer.} These vectors are passed through each layer in order to return the decomposition up to that layer: $\bm{x}^\ell_{i\Leftarrow k} \rightarrow \mathrm{Encoder}^\ell \rightarrow \bm{x}^{\ell+1}_{i\Leftarrow k}$. Ultimately, the decomposed vectors of the \textsc{[CLS]} token are passed through the classification head, which returns a decomposed set of logits. These values reveal the extent to which each token has influenced the corresponding output logit. 

In this section, we explain how vectors are decomposed and propagated through each component, altogether describing a complete propagation through an encoder layer. After this operation is repeated across all layers, we describe how the classification head transforms the decomposition vectors from the last encoder layer into prediction explanation scores.

\subsection{The Multi-head Self-Attention}

The first component in each encoder layer is the multi-head self-attention mechanism. Each head, $h \in \{1,2,...,H\}$, computes a set of attention weights where each weight $\alpha^{h}_{i,j}$ specifies the raw attention from the $i^\text{th}$ to the $j^\text{th}$ token. 
According to \citet{kobayashi-etal-2021-incorporating-residual}'s reformulation, the output of multi-head self-attention, $\bm{z}^{\ell}_{i}$, can be viewed as the sum of the projected value transformation ($\bm{v}^h(\bm{x})=\bm{x}\bm{W}^h_v+\bm{b}^h_v$) of the input over all heads:
\vspace{-1.0ex}
\begin{equation}
\bm{z}^\ell_i=\sum_{h=1}^H\sum_{j=1}^N\alpha^{h}_{i,j}\bm{v}^h(\bm{x}_j^\ell)\bm{W}^h_{\bm{O}}+\bm{b_O}
\end{equation}
The multi-head mixing weight $\bm{W}^h_{\bm{O}}$ and bias $\bm{b_O}$ could be combined with the value transformation to form an equivalent weight $\bm{W}^h_{\bm{Att}}$ and bias $\bm{b_{Att}}$ in a simplified format\footnote{cf. \ref{sec:app_equivalent_att} for further detail on the simplification process.}:
\vspace{-1.0ex}
\begin{equation}
    \bm{z}^\ell_i=\sum_{h=1}^H\sum_{j=1}^N\underbrace{\alpha^{h}_{i,j}\bm{x}_j^\ell \bm{W}^h_{\bm{Att}}}_{\bm{z}_{i\leftarrow j}^\ell} + \bm{b_{Att}}
\end{equation}
Since \citet{kobayashi-etal-2021-incorporating-residual} and \citet{modarressi-etal-2022-globenc} both use local-level decomposition, they regard $\bm{z}_{i\leftarrow j}^\ell$ as the attribution vector of token $i$ from input token $j$ in layer $\ell$'s multi-head attention.\footnote{Note that even though they discard the bias within the head-mixing module, $\bm{b_O}$, the value bias $\bm{b}^h_v$ is included.} We also utilize this attribution vector, but only in the first encoder layer since its inputs are also the same inputs of the whole model ($\bm{z}_{i\leftarrow j}^1 = \bm{z}_{i\Leftarrow j}^1$). For other layers, however, each layer's decomposition should be based on the decomposition of the previous encoder layer. Therefore, we plug Eq. \ref{eq:decomp} into the formula above:
\begin{equation}
\begin{aligned}
    \bm{z}^\ell_i&=\sum_{h=1}^H\sum_{j=1}^N\alpha^{h}_{i,j}\sum_{k=1}^N\bm{x}^{\ell}_{j\Leftarrow k} \bm{W}^h_{\bm{Att}} + \bm{b_{Att}}\\
    &=\sum_{k=1}^N\tcbhighmath{\sum_{h=1}^H\sum_{j=1}^N\alpha^{h}_{i,j}\bm{x}^{\ell}_{j\Leftarrow k} \bm{W}^h_{\bm{Att}}} + \bm{b_{Att}}
    \label{eq:decomp_selfatt}
\end{aligned}
\end{equation}
To finalize the decomposition we need to handle the bias which is outside the model inputs summation ($\sum_{k=1}^N$). One possible workaround would be to simply omit the model's internal biases inside the self-attention layers and other components such as feed-forward networks. We refer to this solution as \emph{NoBias}. However, without the biases, the input summation would be incomplete and cannot recompose the inner representations of the model. Also, if the decomposition is carried out all the way to the classifier's output without considering the biases, the resulting values will not tally up to the logits predicted by the model. To this end, we also introduce a decomposition method for the bias vectors with \emph{AbsDot}, which is based on the absolute value of the dot product of the summation term (highlighted in Eq. \ref{eq:decomp_selfatt}) and the bias:
\begin{equation}
    \omega_k=\frac{|\bm{b_{Att}}\cdot \bm{z}_{i\Leftarrow k,[\mathrm{NoBias}]}^\ell|}{\sum_{k=1}^N|\bm{b_{Att}}\cdot \bm{z}_{i\Leftarrow k,[\mathrm{NoBias}]}^\ell|}
\end{equation}
where $\omega_k$ is the weight that decomposes the bias and enables it to be inside the input summation:
\begin{equation}
\small
\bm{z}^\ell_i=\sum_{k=1}^N\underbrace{ ( \tcbhighmath{\sum_{h=1}^H\sum_{j=1}^N\alpha^{h}_{i,j}\bm{x}^{\ell}_{j\Leftarrow k} \bm{W}^h_{\bm{Att}}} + \omega_k\bm{b_{Att}})}_{\mbox{\normalsize{$\bm{z}_{i\Leftarrow k}^\ell$}}}
\label{eq:decomp_selfatt_w_bias}
\end{equation}
The rationale behind \emph{AbsDot} is that the bias is ultimately added into all vectors at each level; consequently, the most affected decomposed vectors are the ones that have the greatest degree of alignment (in terms of cosine similarity) and also have larger norms. The sole usage of cosine similarity could be one solution but in that case, a decomposed vector lacking a norm (such as padding tokens) could also be affected by the bias vector. Although alternative techniques may be employed, our preliminary quantitative findings suggested that \emph{AbsDot} represents a justifiable and suitable selection.

Our main goal from now on is to try to make the model inputs summation $\sum_{k=1}^N$ the most outer sum, so that the summation term ($\bm{z}_{i\Leftarrow k}^\ell$ for the formula above) ends up as the desired decomposition.\footnote{For a bias-included analysis, note that the bias weighting in all subsequent decomposition equations is always determined by the bias itself and its prior term (highlighted in the above formula).}
\subsection{Finalizing the Attention Module}
After the multi-head attention, a residual connection adds the layer's inputs ($\bm{x}_i^\ell$) to $\bm{z}_i^\ell$, producing the inputs of the first LayerNormalization (LN\#1):
\vspace{-0.5ex}
\begin{equation}
\begin{aligned}
\tilde{\bm{z}}^\ell_i&=\mathrm{LN}(\bm{z^+}^\ell_i)\\
&=\mathrm{LN}(\bm{x}_i^\ell+\sum_{k=1}^N\bm{z}_{i\Leftarrow k}^\ell)\\
&=\mathrm{LN}(\sum_{k=1}^N[\bm{x}^{\ell}_{i\Leftarrow k}+\bm{z}_{i\Leftarrow k}^\ell])
\end{aligned}
\label{eq:decomp_res1}
\vspace{-0.5ex}
\end{equation}
Again, to expand the decomposition over the LN function, we employ a technique introduced by \citet{kobayashi-etal-2021-incorporating-residual} in which the LN function is broken down into a summation of a new function $g(.)$:
\vspace{-1.0ex}
\begin{equation}
\begin{gathered}
    \mathrm{LN}(\bm{z^+}^\ell_i)=\sum_{k=1}^N\underbrace{g_{\bm{z^+}^\ell_i}(\bm{z^+}_{i\Leftarrow k}^\ell)+\bm{\beta}}_{\tilde{\bm{z}}_{i\Leftarrow k}^\ell}\\
    g_{\bm{z^+}^\ell_i}(\bm{z^+}_{i\Leftarrow k}^\ell):=\frac{\bm{z^+}_{i\Leftarrow k}^\ell - m(\bm{z^+}_{i\Leftarrow k}^\ell)}{s(\bm{z^+}^\ell_i)} \odot \bm{\gamma}
\label{eq:decomp_LNDecomp}
\end{gathered}
\end{equation}
where $m(.)$ and $s(.)$ represent the input vector's element-wise mean and standard deviation, respectively.\footnote{$\bm{\gamma} \in \mathbb{R}^d$ and $\bm{\beta} \in \mathbb{R}^d$ are respectively the trainable scaling and bias weights of LN. For extra details, please refer to Appendix A in \citet{kobayashi-etal-2021-incorporating-residual} for the derivation.} Unlike \citet{kobayashi-etal-2021-incorporating-residual} and \citet{modarressi-etal-2022-globenc}, we also include the LN bias ($\bm{\beta}$) using our bias decomposition method.
\subsection{Feed-Forward Networks Decomposition}
\label{sec:FFN_decomp}
Following the attention module, the outputs enter a 2-layer Feed-Forward Network (FFN) with a non-linear activation function ($f_\mathrm{act}$):
\vspace{-1.0ex}
\begin{equation}
\small
\begin{aligned}
    \bm{z}_\mathrm{FFN}^\ell &= \mathrm{FFN}(\tilde{\bm{z}}^\ell_i)\\
    &=f_\mathrm{act}(\underbrace{{\tilde{\bm{z}}^\ell_i} \bm{W}_{\mathrm{FFN}}^1+\bm{b}_{\mathrm{FFN}}^1}_{\mbox{\normalsize{$\bm{\zeta}^\ell_{i}$}}})\bm{W}_{\mathrm{FFN}}^2+\bm{b}_{\mathrm{FFN}}^2
\label{eq:FFN}
\end{aligned}
\vspace{-1.0ex}
\end{equation}
$\bm{W}_{\mathrm{FFN}}^\lambda$ and $\bm{b}_{\mathrm{FFN}}^\lambda$ represent the weights and biases, respectively, with $\lambda$ indicating the corresponding layer within the FFN. In this formulation, the activation function is the primary inhibiting factor to continuing the decomposition. As a workaround, we approximate and decompose the activation function based on two assumptions: the activation function (1) passes through the origin ($f_\mathrm{act}(0)=0$) and (2) is monotonic.\footnote{Even though the \emph{GeLU} activation function, which is commonly used in BERT-based models, is not a monotonic function in its $x<0$ region, we ignore it since the values are small.} The approximate function is simply a zero intercept line with a slope equal to the activation function's output divided by its input in an elementwise manner:
\vspace{-1.0ex}
\begin{equation}
\begin{gathered}
    f^{(\bm{x})}_\mathrm{act}(\bm{x}) = \bm{\theta}^{(\bm{x})} \odot \bm{x} \\
    \bm{\theta}^{(\bm{x})} := (\theta_1, \theta_2, ... \theta_d) \text{ s.t. } \theta_t = \frac{f_\mathrm{act}(x^{(t)})}{x^{(t)}}
\label{eq:FFN_approx}
\end{gathered}
\vspace{-1.0ex}
\end{equation}
where $(t)$ denotes the dimension of the corresponding vector. One important benefit of this alternative function is that when $\bm{x}$ is used as an input, the output is identical to that of the original activation function. Hence, the sum of the decomposition vectors would still produce an accurate result. 
Using the described technique we continue our progress from Eq. \ref{eq:FFN} by decomposing the activation function:
\vspace{-2.0ex}
\begin{equation}
\begin{aligned}
    \bm{z}_{\mathrm{FFN},i}^\ell &=f_\mathrm{act}^{(\bm{\zeta}^\ell_{i})}(\sum_{k=1}^N\bm{\zeta}^\ell_{i\Leftarrow k})\bm{W}_{\mathrm{FFN}}^2+\bm{b}_{\mathrm{FFN}}^2 \\
    &=\sum_{k=1} \underbrace{\bm{\theta}^{(\bm{\zeta}^\ell_{i})} \odot \bm{\zeta}^\ell_{i\Leftarrow k}+\bm{b}_{\mathrm{FFN}}^2}_{\bm{z}_{\mathrm{FFN},i\Leftarrow k}^\ell}
\label{eq:FFN_decomp}
\end{aligned}
\vspace{-1ex}
\end{equation}
In designing this activation function approximation, we prioritized completeness and efficiency. For the former, we ensure that the sum of decomposed vectors should be equal to the token’s representation, which has been fulfilled by applying the same $\theta$ to all decomposed values $\zeta$ based on the line passing the activation point. While more complex methods (such as applying different $\theta$ to each $\zeta$) which require more thorough justification may be able to capture the nuances of different activation functions more accurately, we believe that our approach strikes a good balance between simplicity and effectiveness, as supported by our empirical results.

The final steps to complete the encoder layer progress are to include the other residual connection and LayerNormalization (LN\#2), which could be handled similarly to Eqs. \ref{eq:decomp_res1} and \ref{eq:decomp_LNDecomp}:
\vspace{-2.0ex}
\begin{equation}
\begin{aligned}
    \bm{x}_i^{\ell+1} &= \mathrm{LN}(\sum_{k=1}^N[\underbrace{\tilde{\bm{z}}^\ell_{i\Leftarrow k} + \bm{z}_{\mathrm{FFN},i\Leftarrow k}^\ell}_{\bm{z}_{\mathrm{FFN}^+,i\Leftarrow k}^\ell}]) \\
    &= \sum_{k=1}^N\underbrace{g_{\bm{z}_{\mathrm{FFN}^+,i}^\ell}(\bm{z}_{\mathrm{FFN}^+,i\Leftarrow k}^\ell)+\bm{\beta}}_{\bm{x}_{i\Leftarrow k}^{\ell+1}}\\
\label{eq:encoder_decomp}
\end{aligned}
\vspace{-3.0ex}
\end{equation}
Using the formulations described in this section, we can now obtain $\bm{x}_{i\Leftarrow k}^{\ell+1}$ from $\bm{x}_{i\Leftarrow k}^{\ell}$, and by continuing this process across all layers, $\bm{x}_{i\Leftarrow k}^{L+1}$ is ultimately determined.

\subsection{Classification Head}
\label{sec:clf_head}
Norm- or summation-based vector aggregation could be utilized to convert the decomposition vectors into interpretable attribution scores. However, in this case, the resulting values would only become the attribution of the output token to the input token, without taking into account the task-specific classification head. This is not a suitable representation of the model's decision-making, as any changes to the classification head would have no effect on the vector aggregated attribution scores. 
Unlike previous vector-based methods, we can include the classification head in our analysis thanks to the decomposition propagation described above.\footnote{We also discuss about alternative use cases in section \ref{sec:app_alternative_use}}
As the classification head is also an FFN whose final output representation is the prediction scores $\bm{y}=(y_1, y_2, ..., y_C)$ for each class $c \in {\{1,2,...,C\}}$, we can continue decomposing through this head as well. In general, the \textsc{[CLS]} token representation of the last encoder layer serves as the input for the two-layer (pooler layer + classification layer) classification head:
\begin{equation}
    \bm{y} = u_\mathrm{act}(\bm{x}_{\textsc{[CLS]}}^{L+1}\bm{W}_{\mathrm{pool}} + \bm{b}_{\mathrm{pool}})\bm{W}_{\mathrm{cls}} + \bm{b}_{\mathrm{cls}}
\label{eq:cls_decomp}
\end{equation}
Following the same procedure as in Section \ref{sec:FFN_decomp}, we can now compute the input-based decomposed vectors of the classification head's output $\bm{y}_{k}$ using the decomposition of the \textsc{[CLS]} token, $\bm{x}_{i\Leftarrow k}$. By applying this, in each class we would have an array of attribution scores for each input token, the sum of which would be equal to the prediction score of the model for that class:
\vspace{-1.0ex}
\begin{equation}
\begin{aligned}
    y_c = \sum_{k=1}^N y_{c \Leftarrow k}
\end{aligned}
\vspace{-1.0ex}
\end{equation}
To explain a predicted output, $y_{c \Leftarrow k}$ would be the attribution of the $k^\text{th}$ token to the total prediction score.

\begin{figure*}[t!]
\centering
    \subfloat{
        \includegraphics[width=0.49\textwidth, trim=0 0 0 20, clip] 
        {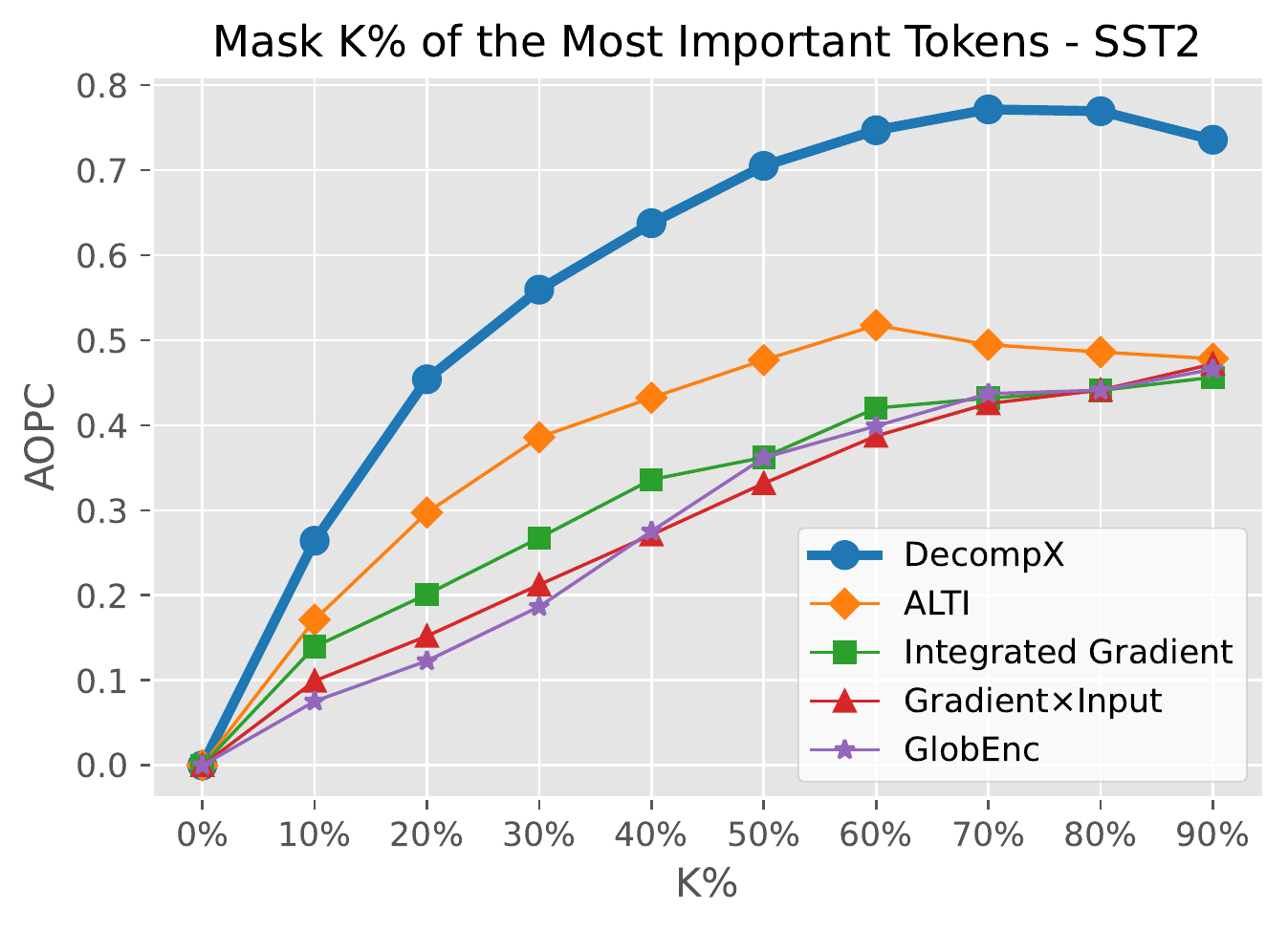}
    }
    \subfloat{
        \includegraphics[width=0.49\textwidth, trim=0 0 0 20, clip] {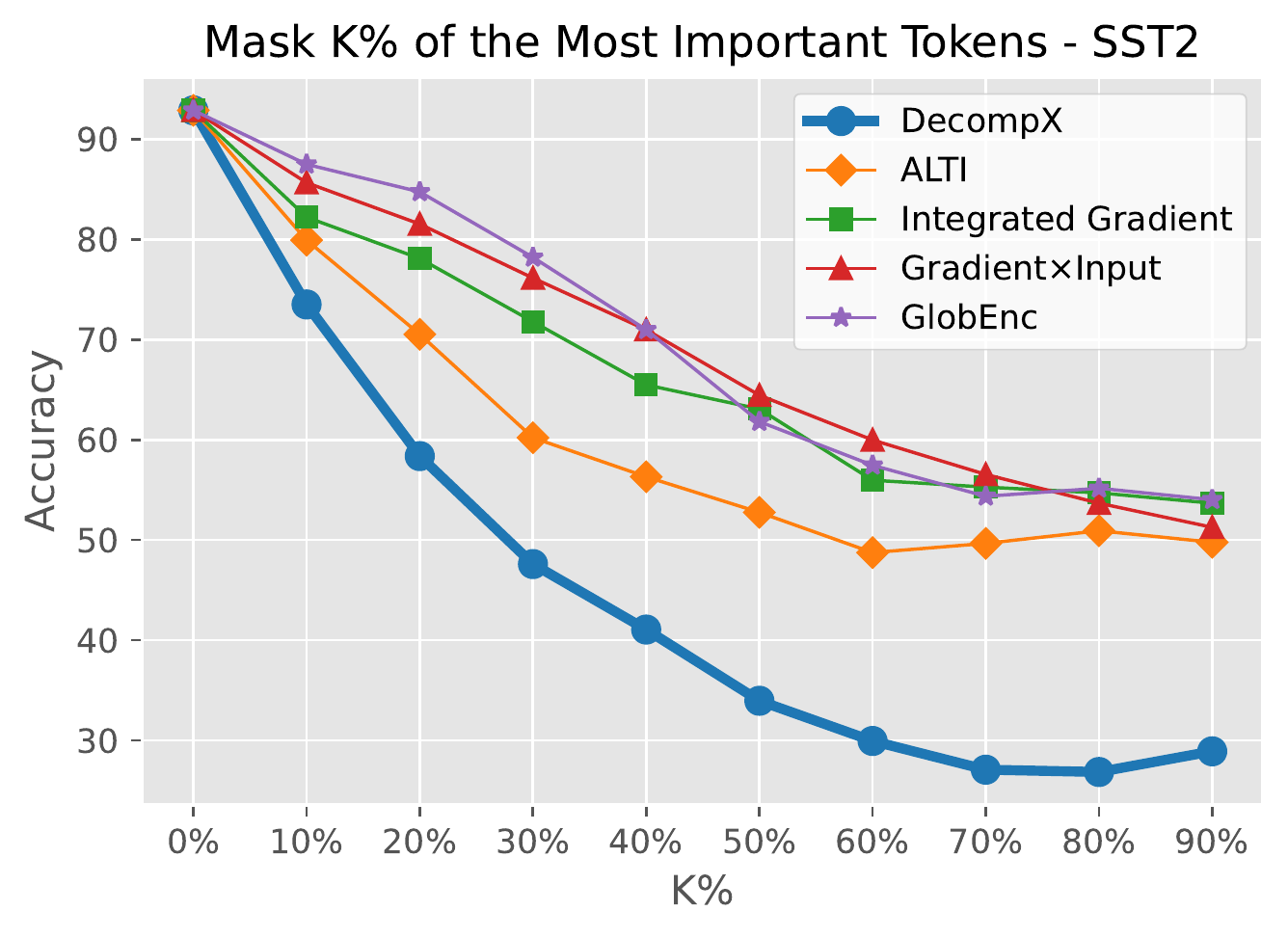}
    }
    \caption{
    AOPC and Accuracy of different explanation methods on SST2 upon masking $K\%$ of the most important tokens (higher AOPC and lower Accuracy are better). DecompX outperforms existing methods by a large margin.
    }
    \label{fig:sst2_max}
\end{figure*}
\begin{table*}[t!]
\begin{center}
\small
\tabcolsep=0.13cm
\begin{tabular}{l c c c | c c c | c c c | c c c} 
 \toprule
     & 
     \multicolumn{3}{c}{\textbf{\textsc{SST2}}} & 
     \multicolumn{3}{c}{\textbf{\textsc{MNLI}}} & 
     \multicolumn{3}{c}{\textbf{\textsc{QNLI}}} & 
     \multicolumn{3}{c}{\textbf{\textsc{HateXplain}}} \\
     \cmidrule(lr){2-13}
    & \scriptsize{\textbf{\textsc{Acc$\downarrow$}}} & \scriptsize{\textbf{\textsc{AOPC$\uparrow$}}} & \scriptsize{\textbf{\textsc{Pred$\uparrow$}}}
     & \scriptsize{\textbf{\textsc{Acc$\downarrow$}}} & \scriptsize{\textbf{\textsc{AOPC$\uparrow$}}} & \scriptsize{\textbf{\textsc{Pred$\uparrow$}}}
     & \scriptsize{\textbf{\textsc{Acc$\downarrow$}}} & \scriptsize{\textbf{\textsc{AOPC$\uparrow$}}} & \scriptsize{\textbf{\textsc{Pred$\uparrow$}}}
     & \scriptsize{\textbf{\textsc{Acc$\downarrow$}}} & \scriptsize{\textbf{\textsc{AOPC$\uparrow$}}} & \scriptsize{\textbf{\textsc{Pred$\uparrow$}}} \\
    \midrule
    GlobEnc \tiny{\cite{modarressi-etal-2022-globenc}}
& 67.14 & 0.307 & 72.36 & 48.07 & 0.498 & 70.43 & 64.93 & 0.342 & 84.00 & 47.65 & 0.401 & 56.50 \\
    \quad + FFN 
& 64.90 & 0.326 & 79.01 & 45.05 & 0.533 & 75.15 & 63.74 & 0.354 & 84.97 & 46.89 & 0.406 & 59.52 \\  
    ALTI \tiny{\cite{ferrando-etal-2022-measuring}}
& 57.65 & 0.416 & 88.30 & 45.89 & 0.515 & 74.24 & 63.85 & 0.355 & 85.69 & 43.30 & 0.469 & 64.67 \\
    \midrule
    Gradient×Input
& 66.69 & 0.310 & 67.20 & 44.21 & 0.544 & 76.05 & 62.93 & 0.366 & 86.27 & 46.28 & 0.433 & 60.67 \\
    Integrated Gradients
& 64.48 & 0.340 & 64.56 & 40.80 & 0.579 & 73.94 & 61.12 & 0.381 & 86.27 & 45.19 & 0.445 & 64.46 \\
    \midrule
    \textbf{DecompX}
& \textbf{40.80}    & \textbf{0.627}           & \textbf{92.20}          & \textbf{32.64}    & \textbf{0.703}         & \textbf{80.95}          & \textbf{57.50}    & \textbf{0.453}         & \textbf{89.84}          & \textbf{38.71}    & \textbf{0.612}         & \textbf{66.34} \\
 \bottomrule
\end{tabular}
\end{center}
\caption{
Accuracy, AOPC, and Prediction Performance of DecompX compared with the existing methods on different datasets. Each figure is the average across all perturbation ratios. As for Accuracy and AOPC, we mask the most important tokens while for Prediction Performance the least important tokens are removed (lower Accuracy, higher AOPC, and higher Prediction Performance scores are better).
}
\label{tab:experiments_max}
\end{table*}

\section{Experiments}
Our faithfulness evaluations are conducted on four datasets covering different tasks, SST-2 \cite{socher-etal-2013-recursive} for sentiment analysis, MNLI \cite{williams-etal-2018-broad} for NLI, QNLI \cite{rajpurkar-etal-2016-squad} for question answering, and HateXplain \cite{mathew2021hatexplain} for hate speech detection.
Our code is implemented based on HuggingFace’s Transformers library \cite{wolf-etal-2020-transformers}. For our experiments, we used fine-tuned BERT-base-uncased \cite{devlin-etal-2019-bert} and RoBERTa-base \cite{liu-roberta-2019}, obtained from the same library.\footnote{RoBERTa results can be found in section \ref{sec:roberta-results}.}
As for gradient-based methods, we choose 0.1 as a step size in integrated gradient experiments and consider the L2-Norm of the token’s gradient vector as its final attribution score.\footnote{All were conducted on an RTX A6000 24GB machine.}

\subsection{Evaluation Metrics}

We aim to evaluate our method's \emph{Faithfulness} by perturbing the input tokens based on our explanations. A widely-used perturbation method removes $K\%$ of tokens with the highest / lowest estimated importance to see its impact on the output of the model \cite{chen-etal-2020-generating-hierarchical, nguyen-2018-comparing}.
To mitigate the consequences of perturbed input becoming out-of-distribution (OOD) for the model, we replace the tokens with [MASK] instead of removing them altogether \cite{deyoung-etal-2020-eraser}. This approach makes the sentences similar to the pre-training data in masked language modeling.
We opted for three metrics: AOPC \cite{samek2016evaluating}, Accuracy \cite{atanasova-etal-2020-diagnostic}, and Prediction Performance \cite{jain-etal-2020-learning}.
\paragraph{AOPC:}
Given the input sentence $x_i$, the perturbed input $\tilde{{x}}^{(K)}_{i}$ is constructed by masking $K\%$ of the most/least important tokens from $x_i$.
Afterward, AOPC computes the average change in the predicted class probability over all test data as follows:
\vspace{-1.0ex}
\begin{equation}
  \label{eq:aopc}
  \text{AOPC}(K)=\frac{1}{N}\sum_{i=1}^N p(\hat{y}\mid{x}_{i})-p(\hat{y}\mid \tilde{{x}}^{(K)}_{i}) 
\vspace{-1.0ex}
\end{equation}
where $N$ is the number of examples, and $p(\hat{y}\mid .)$ is the probability of the predicted class.
When masking the most important tokens, a higher AOPC is better, and vice versa.

\paragraph{Accuracy:}
Accuracy is calculated by averaging the performance of the model over different masking ratios. In cases where tokens are masked in decreasing importance order, lower Accuracy is better, and vice versa.

\paragraph{Predictive Performance:}
\citet{jain-etal-2020-learning} employ predictive performance to assess faithfulness by evaluating the sufficiency of their extracted rationales. The concept of sufficiency evaluates a rationale---a discretized version of soft explanation scores---to see if it adequately indicates the predicted label \cite{jacovi-etal-2018-understanding, yu-etal-2019-rethinking}. Based on this, a BERT-based model is trained and evaluated based on inputs from rationales only to see how it performs compared with the original model. 
As mentioned by \citet{jain-etal-2020-learning}, for each example, we select the top-$K\%$ tokens based on the explanation methods' scores to extract a rationale\footnote{We select the top 20\% for the single sentence and top 40\% for the dual sentence tasks.}.

\subsection{Results}
Figure~\ref{fig:sst2_max} demonstrates the AOPC and Accuracy of the fine-tuned model on the perturbed inputs at different corruption rates $K$. 
As we remove the most important tokens in this experiment, higher changes in the probability of the predicted class computed by AOPC and lower accuracies are better. 
Our method outperforms comparison explanation methods, both vector- and gradient-based, by a large margin at every corruption rate on the SST2 dataset. 
Table~\ref{tab:experiments_max} shows the aggregated AOPC and Accuracy over corruption rates, as well as Predicted Performance on different datasets. 
DecompX consistently outperforms other methods, which confirms that a holistic vector-based approach can present higher-quality explanations.
Additionally, we repeated this experiment by removing the \emph{least} important tokens. Figure~\ref{fig:sst2_min} and Table~\ref{tab:experiments_min} in the Appendix demonstrate that even with 10\%-20\% of the tokens selected by DecompX the task still performs incredibly well. When keeping only 10\% of the tokens based on DecompX, the accuracy only drops by 2.64\% (from 92.89\% of the full sentence), whereas the next best vector- and gradient-based methods suffer from the respective drops of 7.34\% and 15.6\%.
In what follows we elaborate on the reasons behind this superior performance.

\begin{figure}[t]
\centering
    \subfloat{
        \includegraphics[width=0.48\textwidth, trim=20 0 0 0, clip] {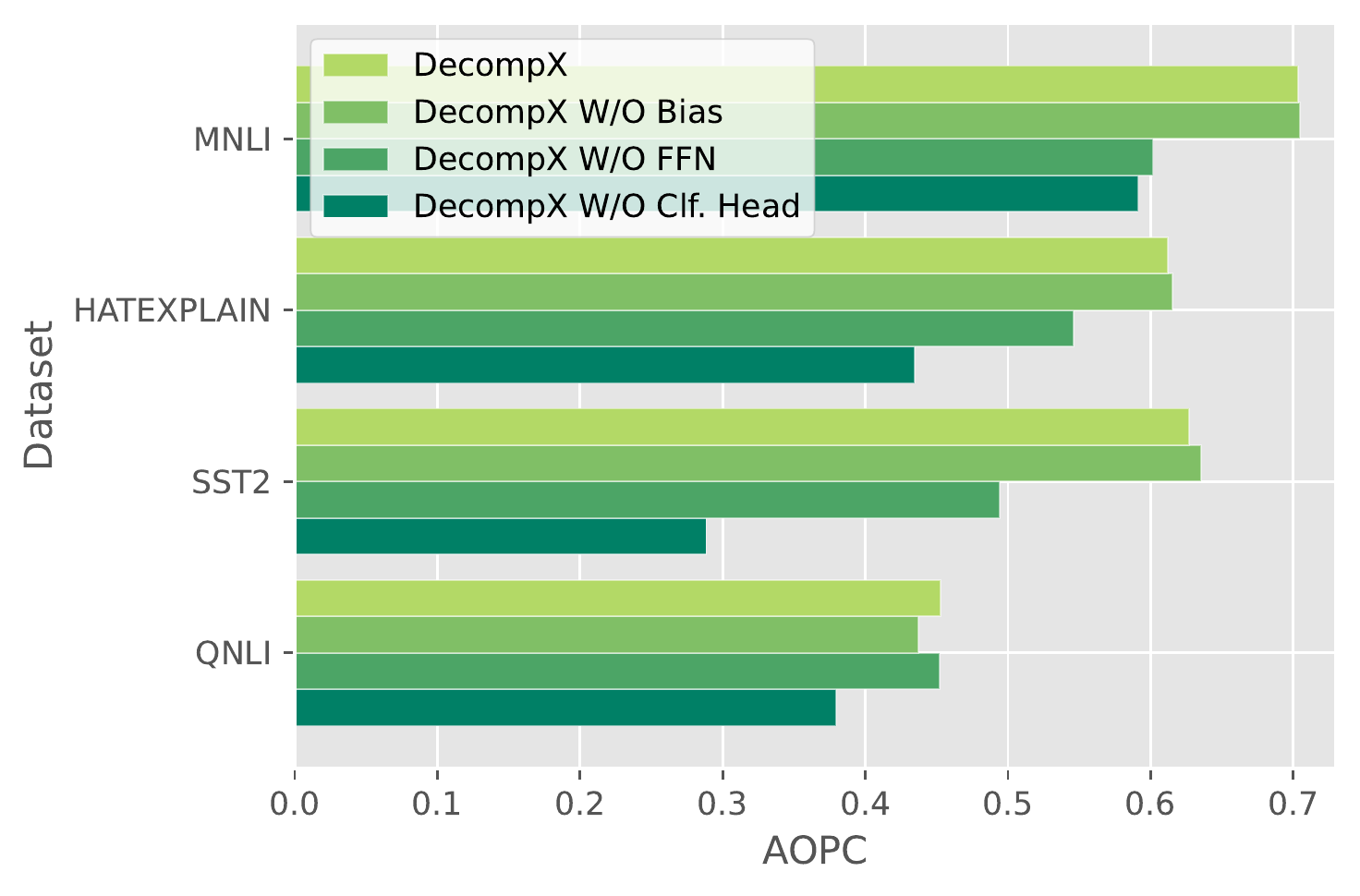}
    }
    \caption{
    Leave-one-out ablation study of DecompX components. Higher AOPC scores are better. 
    }
    \label{fig:ablation_leaveOneOut_max_aopc}
\end{figure}

\paragraph{The role of feed-forward networks.}
Each Transformers encoder layer includes a feed-forward layer. \citet{modarressi-etal-2022-globenc} omitted the influence of FFN when applying decomposition inside each layer due to FFN being a non-linear component. 
In contrast, we incorporated FFN's effect by a point-wise approximation (cf. \S\ref{sec:FFN_decomp}). 
To examine its individual effect we implemented GlobEnc + FFN where we incorporated the FFN component in each layer. Table~\ref{tab:experiments_max} shows that this change improves GlobEnc in terms of faithfulness, bringing it closer to gradient-based methods.
Moreover, we conducted a leave-one-out ablation analysis\footnote{In all our ablation studies, we use norm-based aggregation when not incorporating the classification head: $\Vert \bm{x}^{L+1}_{\textsc{[CLS]}\Leftarrow k}\Vert$}
to ensure FFN's effect on DecompX. Figure~\ref{fig:ablation_leaveOneOut_max_aopc} reveals that removing FFN significantly decreases the AOPC.

\begin{figure*}[t!]
\centering
    \subfloat{
        \includegraphics[width=0.95\textwidth, trim=6 675 0 0, clip] {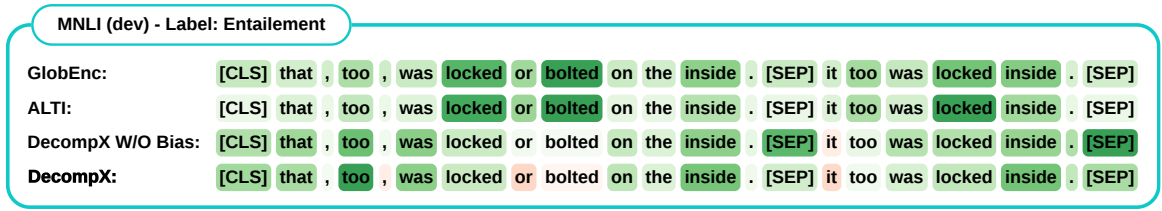}
    }
    \caption{
    An example from MNLI dataset with Entailment label. In DecompX, green/red indicates the positive/negative impact of the token on the predicted label (Entailment, See Figure~\ref{fig:examples_mnli_labels} for Neutral and Contradiction). GlobEnc and ALTI only provide the general importance of tokens, not their positive or negative effect on each output class.
    }
    \label{fig:examples_mnli}
\end{figure*}

\paragraph{The role of biases.}
Even though Figure~\ref{fig:ablation_leaveOneOut_max_aopc} demonstrates that considering bias in the analysis only has a slight effect, it is important to add biases for the human interpretability of DecompX. Figure~\ref{fig:examples_mnli} shows the explanations generated for an instance from MNLI by different methods. While the order of importance is the same in DecompX and DecompX W/O Bias, it is clear that adding the bias fixes the origin and describes which tokens had positive (green) or negative (red) effect on the predicted label probability. 
Another point is that without considering the biases, presumably less influential special tokens such as [SEP] are weighed disproportionately which is corrected in DecompX.\footnote{The importance of special tokens does not change our results as it is not possible to remove the special tokens in the perturbed input.}

\begin{figure}[t]
\centering
    \subfloat{
        \includegraphics[width=0.48\textwidth, trim=20 0 0 0, clip] {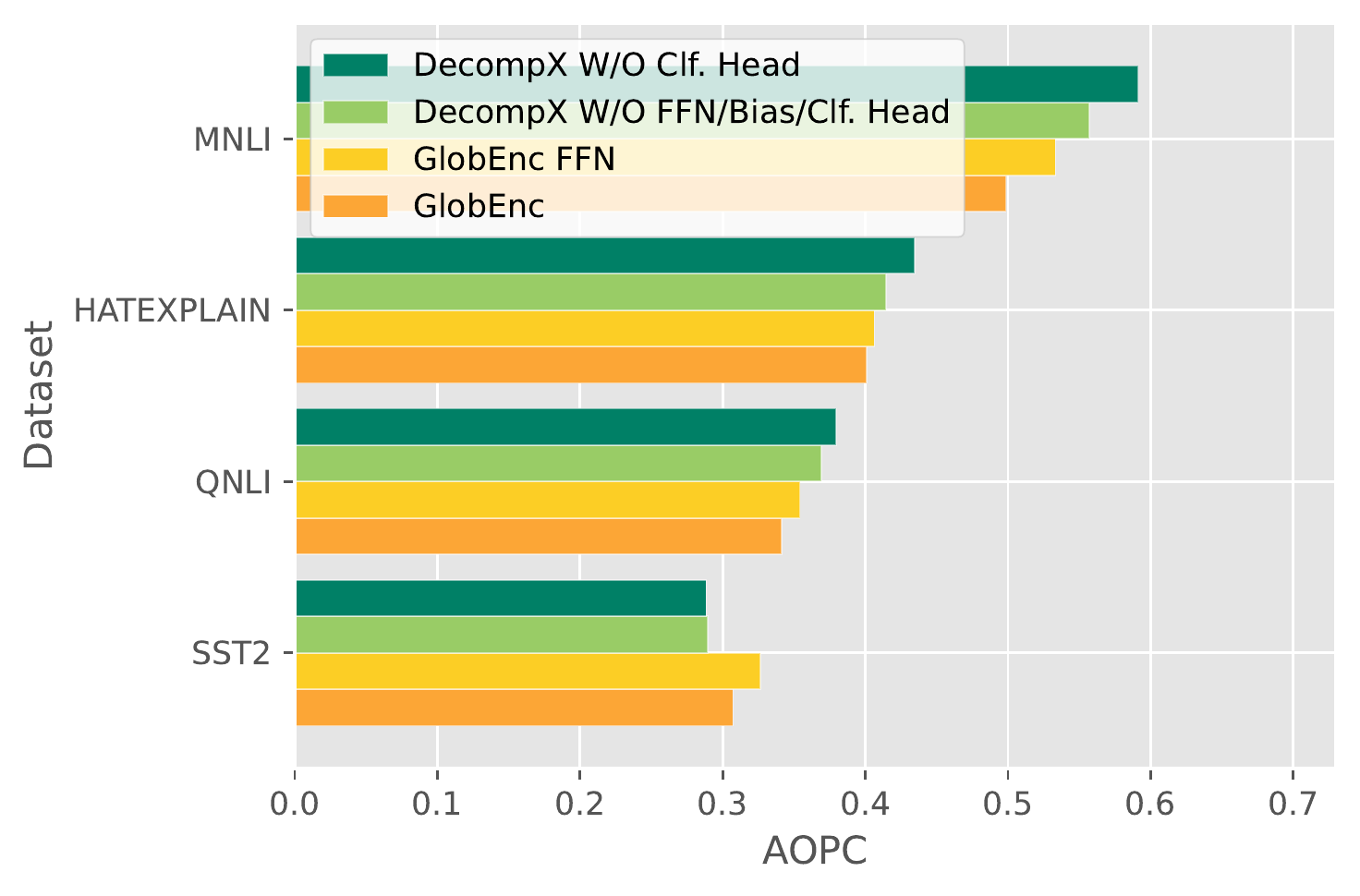}
    }
    \caption{
    Ablation study for illustrating the effect of decomposition. Higher AOPC scores are better. 
    }
    \label{fig:ablation_decomp_max_aopc}
\end{figure}

\paragraph{The role of classification head.}
Figure~\ref{fig:ablation_leaveOneOut_max_aopc} illustrates the effect of incorporating the classification head by removing it from DecompX. AOPC drastically drops when we do not consider the classification head, even more than neglecting bias and FFN, highlighting the important role played by the classification head. Moreover, incorporating the classification head allows us to acquire the exact effect of individual input tokens on each specific output class. 
An example of this was shown earlier in Figure~\ref{fig:examples}, where the explanations are for the predicted class (Positive) in SST2.
Figure~\ref{fig:examples_mnli} provides another example, for an instance from the MNLI dataset.
Due to their omitting of the classification head, previous vector-based methods assign importance to some tokens (such as ``or bolted'') which are actually not important for the predicted label. 
This is due to the fact that the tokens were important for another label (contradiction; cf. Figure~\ref{fig:examples_mnli_labels}).
Importantly, previous methods fall short of capturing this per-label distinction.
Consequently, we believe that no explanation method that omits the classification head can be deemed complete.

\paragraph{The role of decomposition.}
In order to demonstrate the role of propagating the decomposed vectors instead of aggregating them in each layer using rollout, we try to close the gap between DecompX and GlobEnc by simplifying DecompX and incorporating FFN in GlobEnc. 
With this simplification, the difference between DecompX W/O classification head and GlobEnc with FFN setups is that the former propagates the decomposition of vectors while the latter uses norm-based aggregation and rollout between layers. Figure~\ref{fig:ablation_decomp_max_aopc} illustrates the clear positive impact of our decomposition. We show that even without the FFN and bias, decomposition can outperform the rollout-based GlobEnc. These results demonstrate that aggregation in-between layers causes information loss and the final attributions are susceptible to this simplifying assumption.

\section{Conclusions}
In this work, we introduced \emph{DecompX}, an explanation method based on propagating decomposed token vectors up to the classification head, which addresses the major issues of the previous vector-based methods. To achieve this, we incorporated all the encoder layer components including non-linear functions, propagated the decomposed vectors throughout the whole model instead of aggregating them in-between layers, and for the first time, incorporated the classification head resulting in faithful explanations regarding the exact positive or negative impact of each input token on the output classes. Through extensive experiments, we demonstrated that our method is consistently better than existing vector- and gradient-based methods by a wide margin. Our work can open up a new avenue for explaining model behaviors in various situations.
As future work, one can apply the technique to encoder-decoder Transformers, multi-lingual, and Vision Transformers architectures. 

\section*{Limitations}
DecompX is an explanation method for decomposing output tokens based on input tokens of a Transformer model. Although the theory is applicable to other use cases, since our work is focused on English text classification tasks, extra care and evaluation experiments may be required to be used safely in other languages and settings. Due to limited resources, evaluation of large language models such as GPT-2 \cite{radfordlanguage-gpt2} and T5 \cite{raffel-transferlearning-T5} was not viable.

\bibliography{custom}
\bibliographystyle{acl_natbib}


\appendix
\counterwithin{figure}{section}
\counterwithin{table}{section}
\section{Appendix}
\label{sec:appendix}
\subsection{Equivalent Weight and Bias in the Attention Module}
\label{sec:app_equivalent_att}
\begin{equation}
\begin{aligned}
    \bm{z}^\ell_i&=\sum_{h=1}^H\sum_{j=1}^N\alpha^{h}_{i,j}(\bm{x}_j^\ell\bm{W}^h_v+\bm{b}^h_v)\bm{W}^h_{\bm{O}}+\bm{b_O}\\
    &=\sum_{h=1}^H\sum_{j=1}^N\alpha^{h}_{i,j}(\bm{x}_j^\ell\bm{W}^h_v\bm{W}^h_{\bm{O}}+\bm{b}^h_v\bm{W}^h_{\bm{O}})+\bm{b_O}\\
    &=\sum_{h=1}^H\sum_{j=1}^N\alpha^{h}_{i,j}\bm{x}_j^\ell \underbrace{\bm{W}^h_v\bm{W}^h_{\bm{O}}}_{\bm{W}^h_{\bm{Att}}}\\
    &\quad+\underbrace{\sum_{h=1}^H\bm{b}^h_v\bm{W}^h_{\bm{O}}\cancelto{1}{\sum_{j=1}^N\alpha^{h}_{i,j}}+\bm{b_O}}_{\bm{b_{Att}}}
\end{aligned}
\end{equation}

\subsection{Alternative use cases}
\label{sec:app_alternative_use}
The versatility of DecompX allows for explaining various NLP tasks and use cases. Since each output representation is decomposed based on the inputs ($\bm{x}^{L+1}_{i\Leftarrow k}$), it can be propagated through the task-specific head. In Question Answering (QA), for instance, there are two heads to identify the beginning and end of the answer span \cite{devlin-etal-2019-bert}. Thanks to the fact that DecompX is applied post-hoc and the final predicted span is known ($\bm{x}^{L+1}_{i=\mathrm{Start}}$ and $\bm{x}^{L+1}_{i=\mathrm{End}}$), we can continue propagation through the heads as described in Section \ref{sec:clf_head}. In the end, DecompX can indicate the impact of each input token on the span selection: $\bm{y}_{\mathrm{Start}\Leftarrow k} \in \mathbb{R}^N$ \& $\bm{y}_{\mathrm{End}\Leftarrow k} \in \mathbb{R}^N$.

\begin{figure*}[t!]
\centering
    \subfloat{
        \includegraphics[width=0.95\textwidth, trim=5 690 0 0, clip] {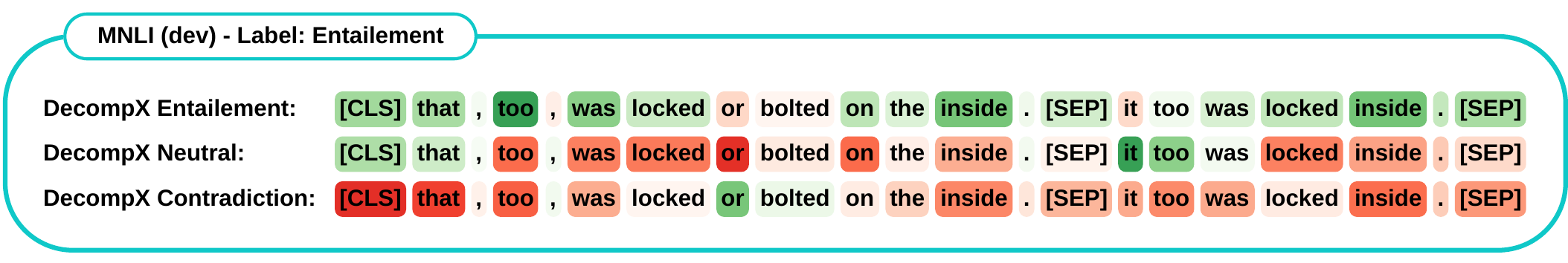}
    }
    \caption{
    An example from MNLI dataset with the \textit{entailment} label. DecompX can provide explanations for each output class, and the sum of input explanations is equal to the final predicted logit for the corresponding class.
    }
    \label{fig:examples_mnli_labels}
\end{figure*}

\subsection{RoBERTa Results}
\label{sec:roberta-results}
\begin{figure*}[t!]
\centering
    \subfloat{
        \includegraphics[width=0.41\textwidth, trim=0 0 0 0, clip] 
        {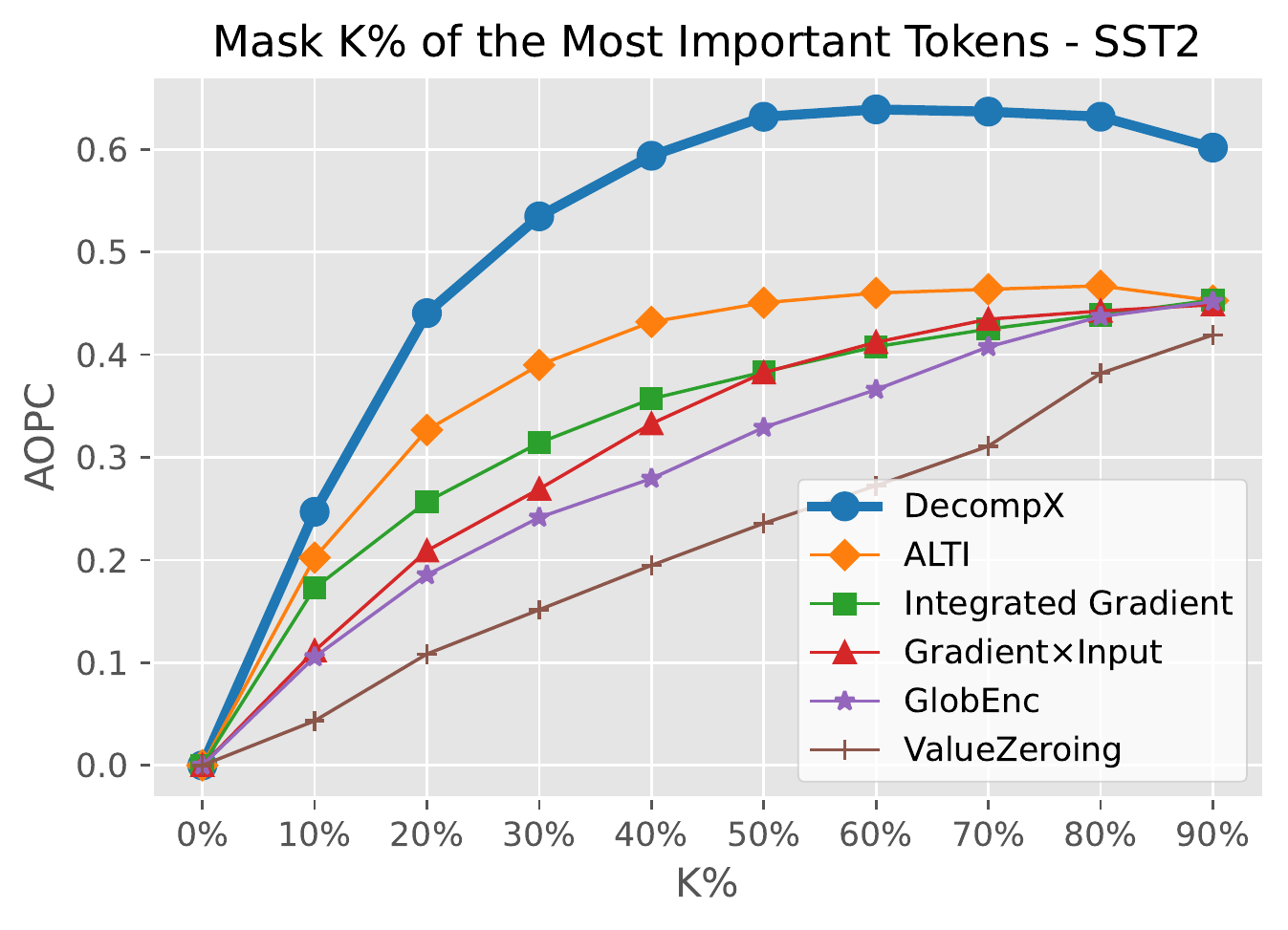}
    }
    \subfloat{
        \includegraphics[width=0.41\textwidth, trim=0 0 0 0, clip] {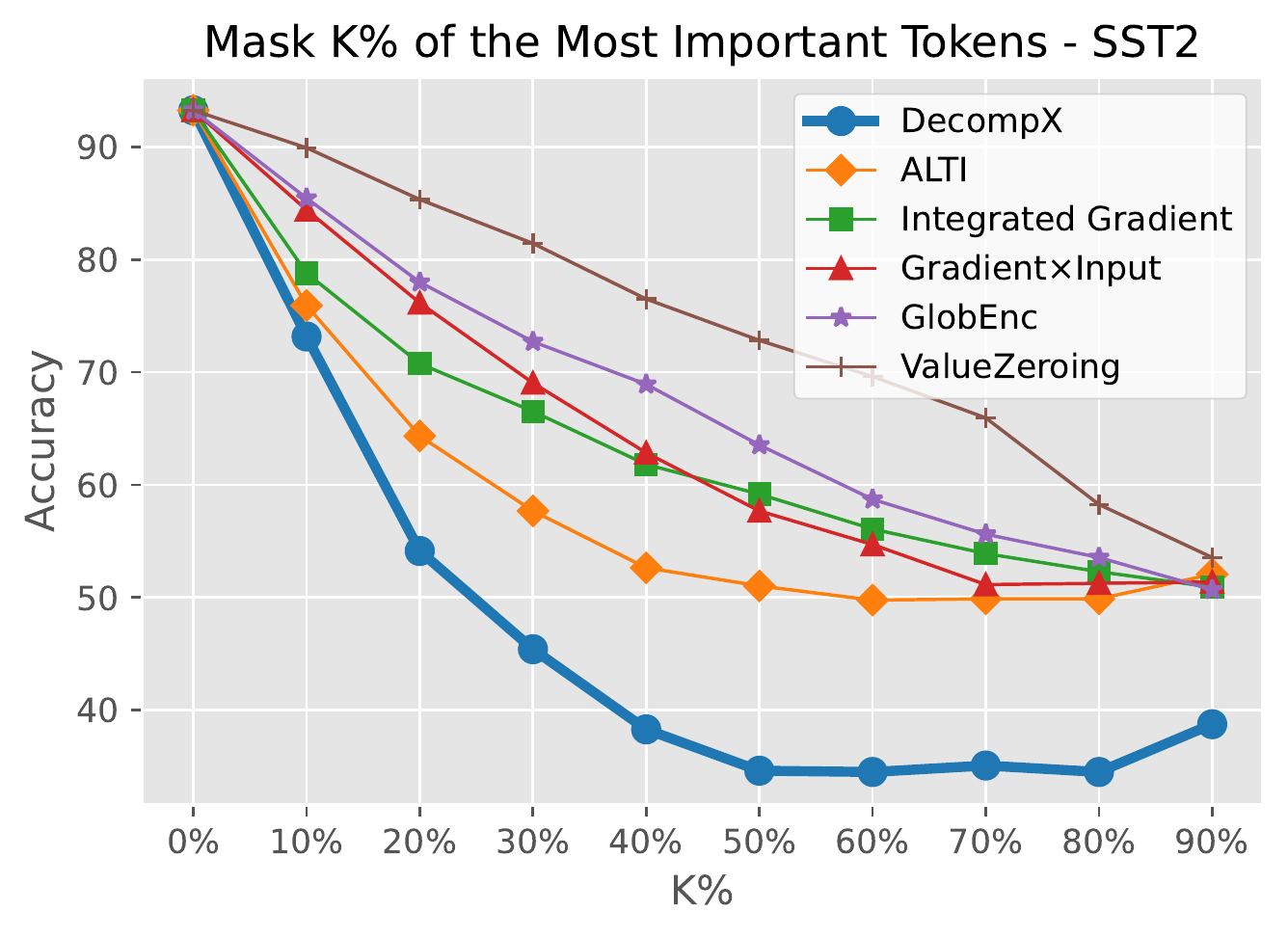}
    }
    \caption{
    RoBERTa-base AOPC and Accuracy of different explanation methods on the SST2 dataset after masking $K\%$ of the most important tokens (higher AOPC and lower Accuracy scores are better).
    }
    \label{fig:sst2_max_roberta}
\end{figure*}

\begin{figure*}[t!]
\centering
    \subfloat{
        \includegraphics[width=0.41\textwidth, trim=0 0 0 0, clip] 
        {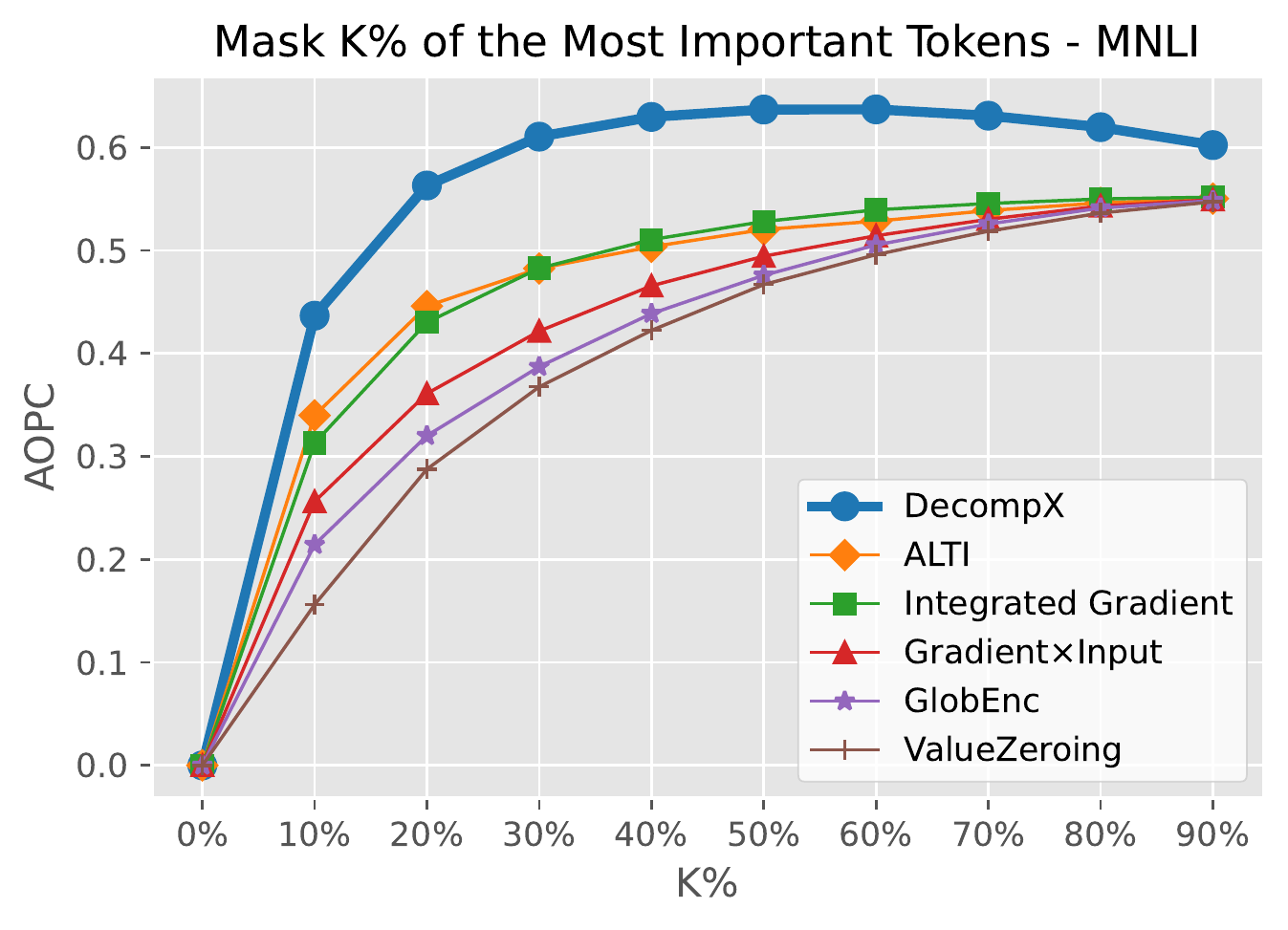}
    }
    \subfloat{
        \includegraphics[width=0.41\textwidth, trim=0 0 0 0, clip] {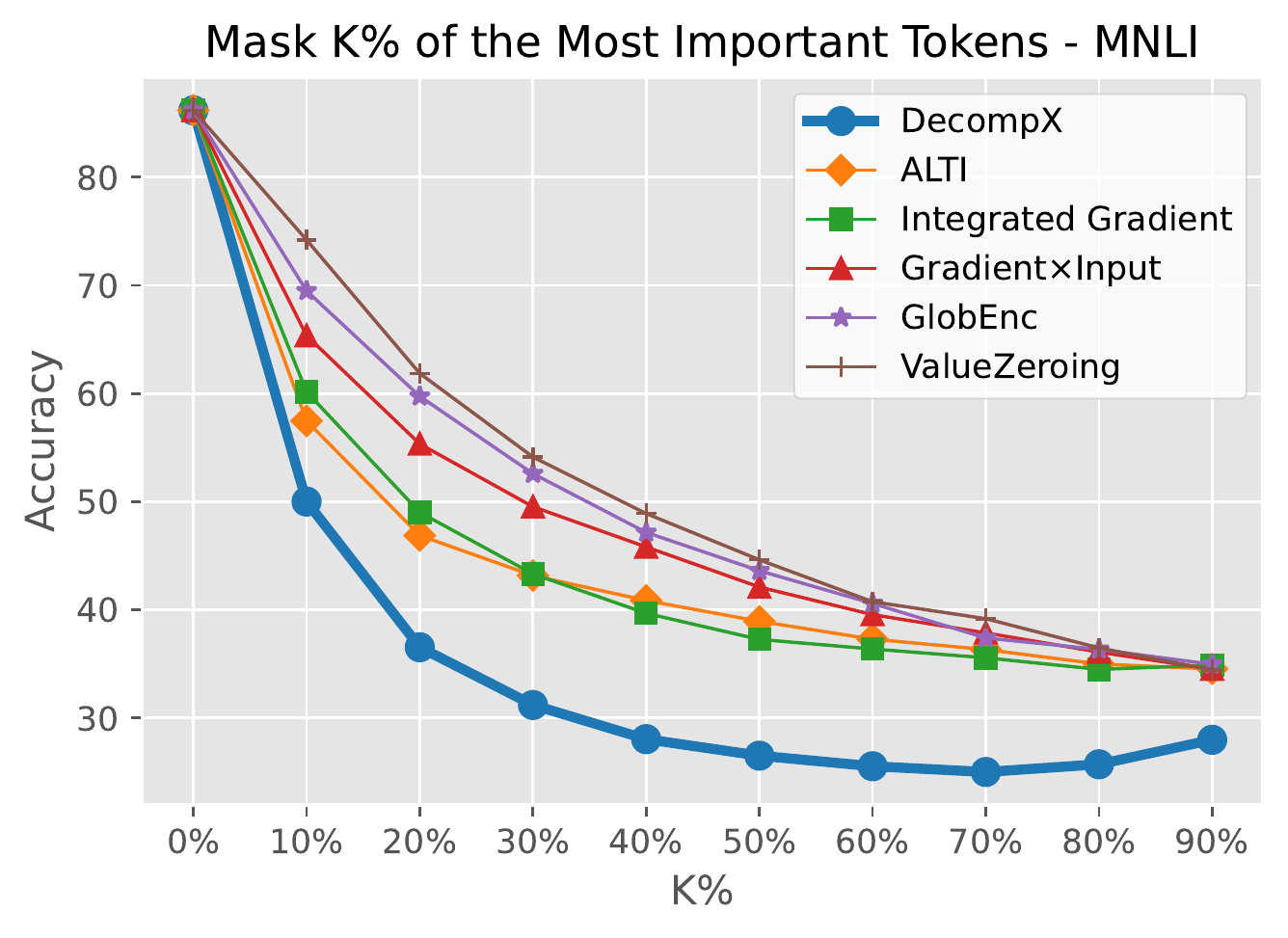}
    }
    \caption{
    RoBERTa-base AOPC and Accuracy of different explanation methods on the MNLI dataset after masking $K\%$ of the most important tokens (higher AOPC and lower Accuracy scores are better).
    }
    \label{fig:mnli_max_roberta}
\end{figure*}

Figures~\ref{fig:sst2_max_roberta} and \ref{fig:mnli_max_roberta} demonstrate the results of our evaluations over the RoBERTa-base model.

In a contemporaneous work, \citet{mohebbi-etal-2023-quantifying} introduced the concept of \emph{ValueZeroing} to incorporate the entire encoder layer and compute context mixing scores in each layer. 
Our experiments, as shown in Figures~\ref{fig:sst2_max_roberta} and \ref{fig:mnli_max_roberta}, demonstrate the poor performance of this technique at global-level.
While it's possible that mismatching configurations\footnote{The authors of the study evaluated the models using blimp probing tasks in a prompting format, whereas we fine-tuned our models on SST-2 and MNLI tasks.} contributed to this inconsistency, we believe that the main issue lies in their reliance on an oversimplified evaluation measure for their global-level assessments.
Their global level evaluation is based on the Spearman's correlation between the blank-out scores and various attribution methods (see Section~7 in \citet{mohebbi-etal-2023-quantifying}).
The issue with this evaluation is that the blank-out baseline scores were obtained by removing only one token from the input (leave-one-out) and measuring the change in prediction probability, which cannot capture feature interactions \cite{towards-faithful-survey-2022}. For instance, in the sentence ``The movie was great and amusing'', independently removing ``great'' or ``amusing'' may not change the sentiment, resulting in smaller scores for these words.

\begin{figure*}[t!]
\centering
    \subfloat{
        \includegraphics[width=0.41\textwidth, trim=0 0 0 0, clip] {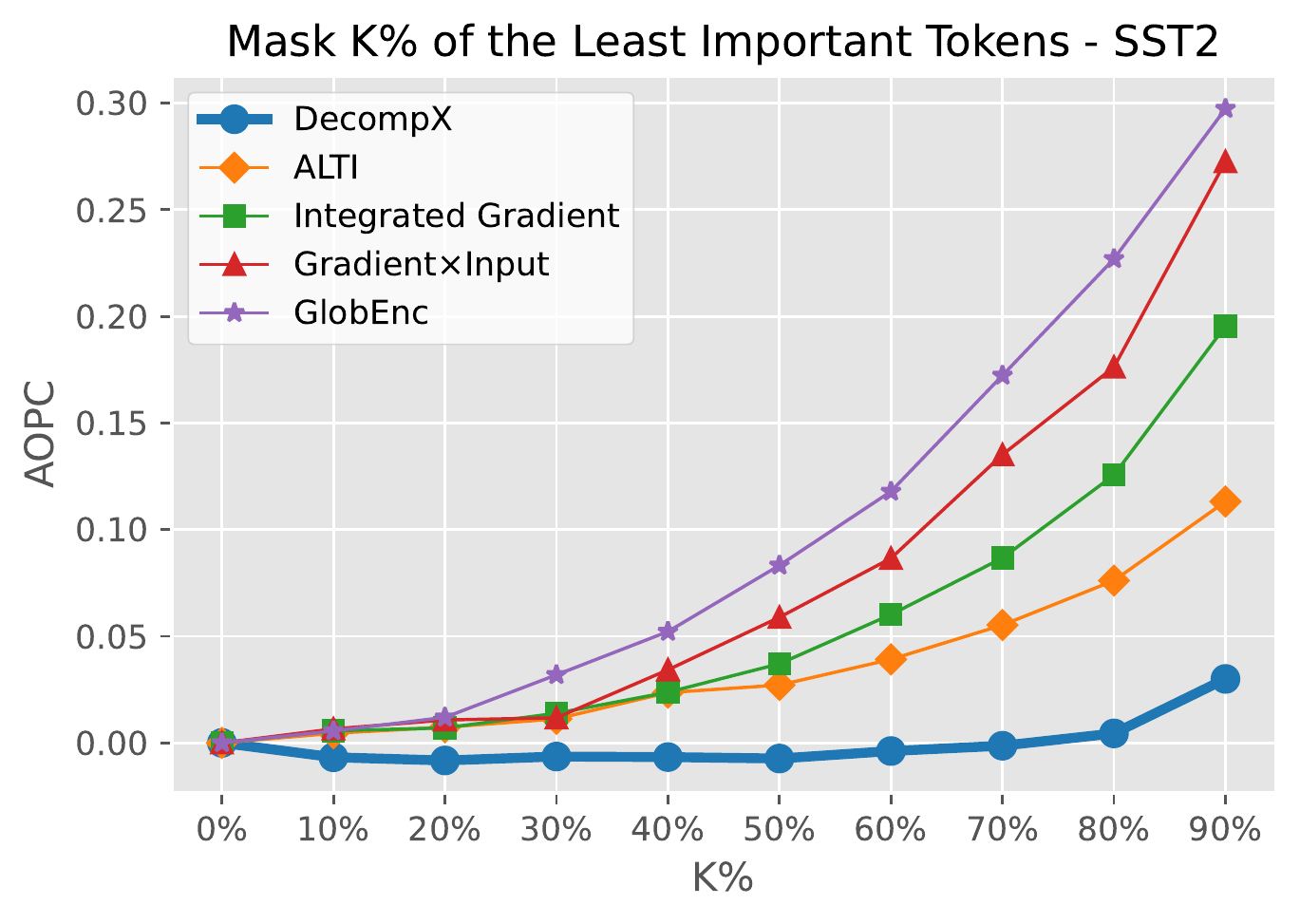}
    }
    \subfloat{
        \includegraphics[width=0.41\textwidth, trim=0 0 0 0, clip] {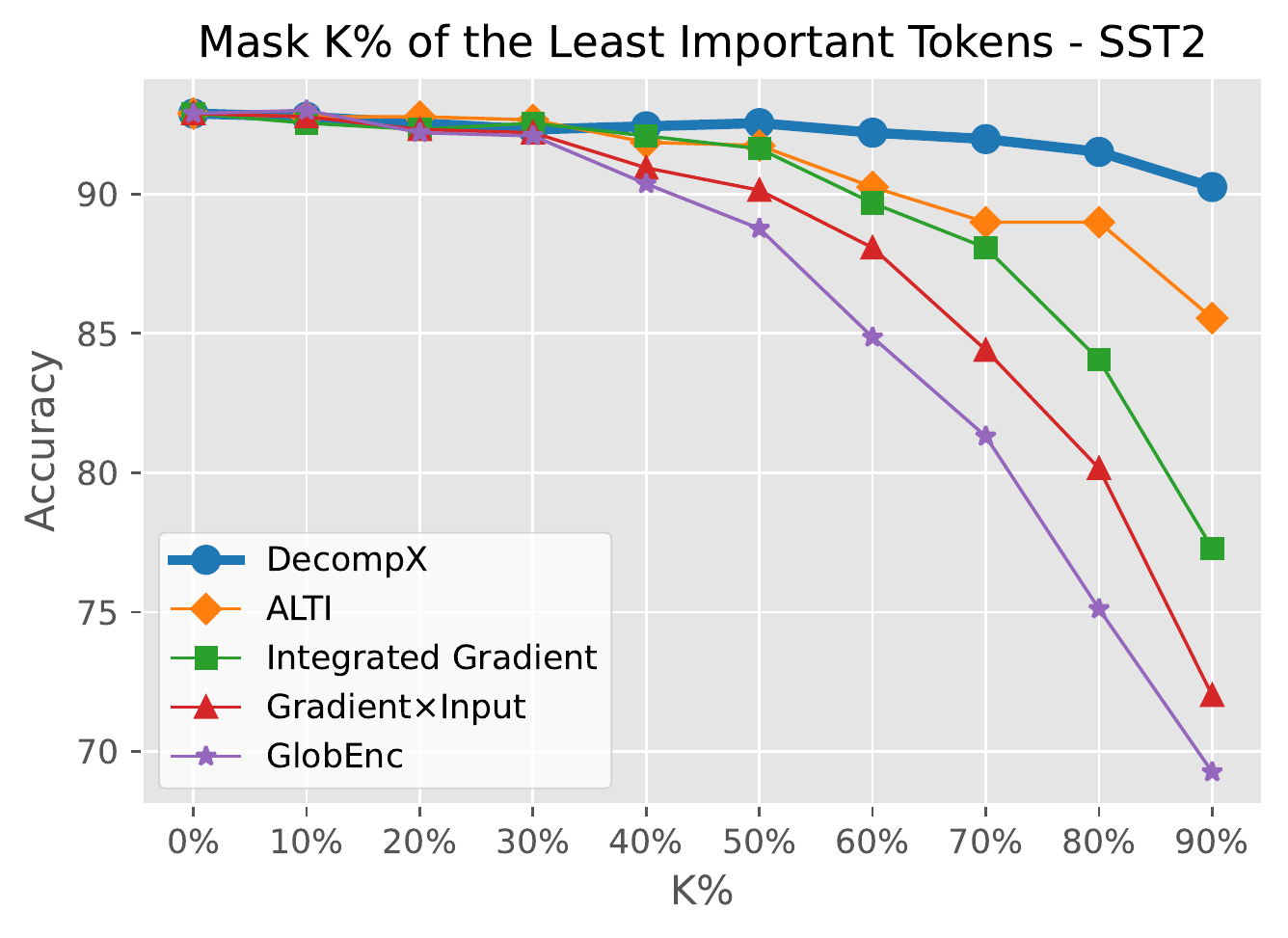}
    }
    \caption{
    AOPC and Accuracy of different explanation methods on the SST2 dataset after masking $K\%$ of the \emph{least} important tokens (lower AOPC and higher Accuracy scores are better).
    }
    \label{fig:sst2_min}
\end{figure*}

\begin{table*}[t!]
\begin{center}
\resizebox{0.97\textwidth}{!}{
\setlength{\tabcolsep}{9pt}
\begin{tabular}{l c c | c c | c c | c c} 
 \toprule
     & 
     \multicolumn{2}{c}{\textbf{\textsc{SST2}}} & 
     \multicolumn{2}{c}{\textbf{\textsc{MNLI}}} & 
     \multicolumn{2}{c}{\textbf{\textsc{QNLI}}} & 
     \multicolumn{2}{c}{\textbf{\textsc{HateXplain}}} \\
     \cmidrule(lr){2-9}
     & \scriptsize{\textbf{\textsc{AOPC$\uparrow$}}} & \scriptsize{\textbf{\textsc{Acc$\downarrow$}}}
     & \scriptsize{\textbf{\textsc{AOPC$\uparrow$}}} & \scriptsize{\textbf{\textsc{Acc$\downarrow$}}}
     & \scriptsize{\textbf{\textsc{AOPC$\uparrow$}}} & \scriptsize{\textbf{\textsc{Acc$\downarrow$}}}
     & \scriptsize{\textbf{\textsc{AOPC$\uparrow$}}} & \scriptsize{\textbf{\textsc{Acc$\downarrow$}}} \\
    \midrule

DecompX & \underline{0.627} & \underline{40.80} & \underline{0.703} & \underline{32.64} & \textbf{0.453} & \underline{57.50} & \underline{0.612}& \textbf{38.71} \\
\quad w/o Bias                & \textbf{0.635}          & \textbf{39.95}           & \textbf{0.705}          & \textbf{32.55}           & 0.437                   & 58.66                    & \textbf{0.615}          & \underline{38.73}             \\
\quad w/o FFN                 & 0.494                   & 53.05                    & 0.601                   & 40.22                    & \underline{0.452}           & \textbf{55.97}           & 0.546                   & 41.24                     \\
\quad w/o Classification Head & 0.288                   & 69.93                    & 0.591                   & 39.80                    & 0.380                   & 61.83                    & 0.435                   & 45.31                     \\
 \bottomrule
\end{tabular}
}
\end{center}
\captionsetup{aboveskip=0pt}
\caption{
Complete results of our ablation study when masking the \emph{most} important tokens. We employ Leave-one-out ablation analysis to demonstrate the effects of bias, FFN, and classification head on the faithfulness of our method.
}
\label{tab:ablation_max_full}
\end{table*}

\begin{table*}[t!]
\begin{center}
\tabcolsep=0.25cm
\resizebox{0.97\textwidth}{!}{
\begin{tabular}{l c c | c c | c c | c c} 
 \toprule
     & 
     \multicolumn{2}{c}{\textbf{\textsc{SST2}}} & 
     \multicolumn{2}{c}{\textbf{\textsc{MNLI}}} & 
     \multicolumn{2}{c}{\textbf{\textsc{QNLI}}} & 
     \multicolumn{2}{c}{\textbf{\textsc{HateXplain}}} \\
     \cmidrule(lr){2-9}
     & \scriptsize{\textbf{\textsc{AOPC$\downarrow$}}} & \scriptsize{\textbf{\textsc{Acc$\uparrow$}}}
     & \scriptsize{\textbf{\textsc{AOPC$\downarrow$}}} & \scriptsize{\textbf{\textsc{Acc$\uparrow$}}}
     & \scriptsize{\textbf{\textsc{AOPC$\downarrow$}}} & \scriptsize{\textbf{\textsc{Acc$\uparrow$}}}
     & \scriptsize{\textbf{\textsc{AOPC$\downarrow$}}} & \scriptsize{\textbf{\textsc{Acc$\uparrow$}}} \\
    \midrule
    GlobEnc \tiny{\cite{modarressi-etal-2022-globenc}}
& 0.111 & 0.852 & 0.205 & 0.715 & 0.151 & 0.817 & 0.204 & 0.600 \\
    \quad + FFN 
& 0.087 & 0.872 & 0.171 & 0.744 & 0.134 & 0.832 & 0.185 & 0.613 \\  
    ALTI \tiny{\cite{ferrando-etal-2022-measuring}}
& 0.040 & 0.906 & 0.191 & 0.731 & 0.121 & 0.844 & 0.135 & 0.644 \\
    \midrule
    Gradient×Input
& 0.088 & 0.870 & 0.164 & 0.746 & 0.125 & 0.839 & 0.175 & 0.620 \\
    Integrated Gradients
& 0.062 & 0.889 & 0.203 & 0.705 & 0.127 & 0.837 & 0.156 & 0.635 \\
    \midrule
    \textbf{DecompX}
& \textbf{-0.001} & \textbf{0.921} & \textbf{0.104} & \textbf{0.767} & \textbf{0.085} & \textbf{0.853} & \textbf{0.035} & \textbf{0.657} \\
 \bottomrule
\end{tabular}
}
\end{center}
\captionsetup{aboveskip=0pt}
\caption{
AOPC and Accuracy of DecompX compared with existing methods on different datasets. AOPC and Accuracy are the averages over perturbation ratios while masking the \emph{least} important tokens (lower AOPC and higher Accuracy are better).
}
\label{tab:experiments_min}
\end{table*}

\end{document}